\documentclass[10pt]{article}

\usepackage[margin=1in]{geometry}
\usepackage{amsmath, amssymb}
\usepackage{graphicx}
\usepackage{booktabs}
\usepackage{hyperref}
\usepackage{cite}
\usepackage{xcolor}
\definecolor{basecompute}{HTML}{157FA2}
\usepackage{microtype}
\usepackage{listings}
\usepackage{caption}
\usepackage{subcaption}
\usepackage{url}
\usepackage{xurl}
\usepackage[numbers,sort&compress]{natbib}
\usepackage{multirow}
\usepackage{placeins}
\usepackage{pgfplots}
\pgfplotsset{compat=1.18}
\usepgfplotslibrary{groupplots}

\title{%
  \vspace{-3em}
  \makebox[\textwidth][l]{\includegraphics[width=0.25\textwidth]{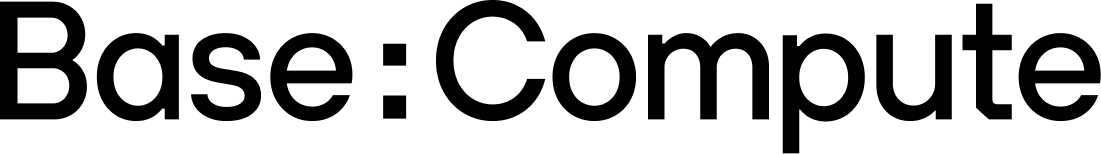}}\\[-1.1em]
  {\color{gray}\rule{1.0\textwidth}{0.4pt}}\\[3em]
  \textbf{BaseRT: Best-in-Class LLM Inference\\ on Apple Silicon via Native Metal}}

\author{
  Prabod Rathnayaka\thanks{Correspondence: \href{mailto:prabod@basecompute.co}{prabod@basecompute.co}} \quad Fabian Waschkowski \quad Lukas Wesemann \\[8pt]
  \textit{Base Compute, Melbourne, Australia} \\
}

\date{}

\begin{document}

\maketitle

\begin{abstract}
We present BaseRT, a native Metal inference runtime for large language models (LLMs) on Apple Silicon, and report the highest inference throughput on this hardware to date. Existing runtimes, including llama.cpp and MLX-based frameworks, incur overhead from abstractions not designed for Metal's execution model or Apple Silicon's unified memory topology. By building natively on Metal with chip-specific kernel fusion, unified memory-aware optimisation, and custom dispatch logic, BaseRT recovers performance that framework-based approaches leave on the table. BaseRT supports a wide range of model families across eight quantisation formats (Q2 to FP16) on all Apple M-series devices. In this paper, we evaluate the Qwen3, Llama 3.2, and Gemma 4 families at Q4 and Q8 quantisation on M3 and M4 Pro devices. BaseRT achieves up to $1.56\times$ higher decode throughput than llama.cpp and up to $1.35\times$ higher than MLX, with substantially larger margins on prefill for mixture-of-experts models, delivering consistent best-in-class throughput from sub-1B to 30B parameter models. These results establish Apple Silicon as a more capable inference platform than previously reported, with direct implications for the emerging edge inference paradigm: as privacy requirements, latency constraints, and cloud cost pressures drive inference toward on-device deployment, performance-optimised local runtimes are a critical enabling layer for this transition. BaseRT is publicly available at \url{https://github.com/basecompute/baseRT}.
\end{abstract}

\section{Introduction}

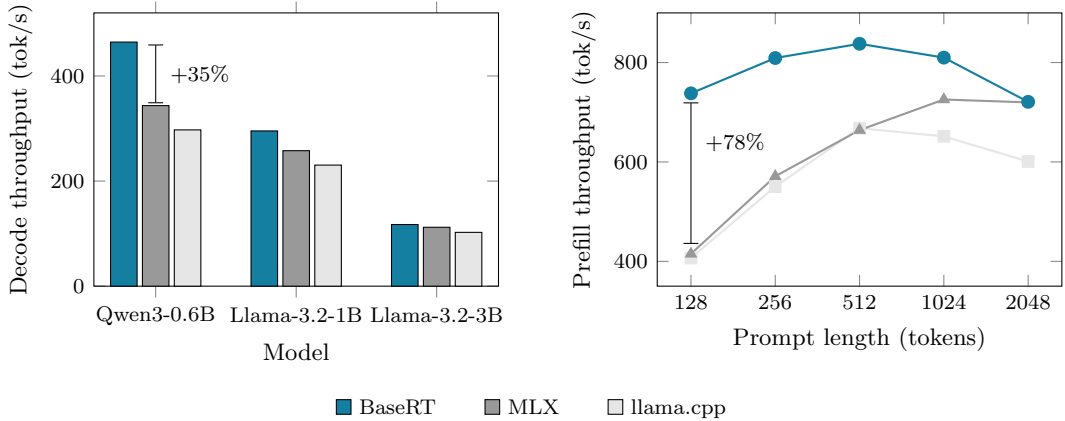
\begin{figure}[t]
  \centering
  \begin{tikzpicture}
    \begin{groupplot}[
      group style={group size=2 by 1, horizontal sep=2.1cm},
      width=0.42\textwidth, height=5.2cm,
      xlabel style={font=\small},
      ylabel style={font=\small},
      xticklabel style={font=\footnotesize},
      yticklabel style={font=\footnotesize},
    ]
    \nextgroupplot[
      ybar, bar width=10pt,
      symbolic x coords={Qwen3-0.6B, Llama-3.2-1B, Llama-3.2-3B},
      xtick=data,
      enlarge x limits=0.22,
      ymin=0, ymax=520,
      xlabel={Model},
      ylabel={Decode throughput (tok/s)},
    ]
      \addplot[fill=basecompute, draw=black] coordinates {
        (Qwen3-0.6B,464.5) (Llama-3.2-1B,295.4) (Llama-3.2-3B,117.3)};
      \addplot[fill=black!40, draw=black] coordinates {
        (Qwen3-0.6B,343.6) (Llama-3.2-1B,257.8) (Llama-3.2-3B,112.1)};
      \addplot[fill=gray!20, draw=black] coordinates {
        (Qwen3-0.6B,297.4) (Llama-3.2-1B,230.4) (Llama-3.2-3B,102.4)};
      \draw[black, |-|] (axis cs:Qwen3-0.6B,348) -- (axis cs:Qwen3-0.6B,460);
      \node[anchor=west, black, font=\footnotesize, xshift=2pt]
        at (axis cs:Qwen3-0.6B,404) {$+35\%$};
    \nextgroupplot[
      xmode=log, log basis x=2,
      xtick={128,256,512,1024,2048},
      xticklabels={128,256,512,1024,2048},
      xlabel={Prompt length (tokens)},
      ylabel={Prefill throughput (tok/s)},
      ymin=350, ymax=900,
      mark options={scale=1.1, fill opacity=1},
      every axis plot/.append style={thick},
    ]
      \addplot[gray!20, mark=square*, mark options={fill=gray!20}] coordinates {
        (128,406.7) (256,550.4) (512,668.0) (1024,651.5) (2048,600.9)};
      \addplot[black!40, mark=triangle*, mark options={fill=black!40}] coordinates {
        (128,415.0) (256,571.1) (512,664.1) (1024,725.7) (2048,719.9)};
      \addplot[basecompute, mark=*] coordinates {
        (128,738.1) (256,809.2) (512,837.9) (1024,810.0) (2048,721.0)};
      \draw[black, |-|] (axis cs:128,435.0) -- (axis cs:128,720);
      \node[anchor=west, black, font=\footnotesize, xshift=2pt]
        at (axis cs:128,640) {$+78\%$};
    \end{groupplot}
  \end{tikzpicture}
  \par\vspace{0.5em}
  {\footnotesize
    \tikz[baseline=-0.8ex]{\filldraw[fill=basecompute,draw=black] (0,-0.8ex) rectangle (1.6ex,0.8ex);}~BaseRT\hspace{1.8em}
    \tikz[baseline=-0.8ex]{\filldraw[fill=black!40,draw=black] (0,-0.8ex) rectangle (1.6ex,0.8ex);}~MLX\hspace{1.8em}
    \tikz[baseline=-0.8ex]{\filldraw[fill=gray!20,draw=black] (0,-0.8ex) rectangle (1.6ex,0.8ex);}~llama.cpp}
  \caption{BaseRT delivers best-in-class throughput on Apple Silicon. Decode throughput (left) is improved by $1.15$--$1.56\times$ over llama.cpp and up to $1.35\times$ over MLX across the Q4 models tested on Apple M4 Pro. For prefill throughput (right), BaseRT leads across all prompt lengths for the Qwen3-30B-A3B model in 4-bit quantisation with up to $1.78\times$ improvement over MLX.}
  \label{fig:hero}
\end{figure}

AI inference demand is growing at a rate that challenges the scalability of centralised cloud infrastructure. Token usage across major providers has grown by roughly an order of magnitude year-on-year as evidenced by token usage via Google Cloud API and a broader mix of providers on Open Router ~\cite{alphabet2025q4, openrouter2026rankings}. However, the majority of enterprise adoption is still nascent and this trend is likely to continue~\cite{mckinsey2025stateofai}. Global inference compute is projected to overtake training compute for the first in 2027, with AI inference expected to make up $40\%$ of global data center compute by 2030 \cite{mckinsey2025aiworkloads}. This trajectory places increasing pressure on the economic and architectural assumptions underlying cloud-only inference, and makes a compelling case for a structural shift toward edge inference: running LLMs locally on consumer and professional hardware. The case for this shift rests on four independent drivers.\\ \\
\textit{Privacy and Data Residency}: Prompts processed in the cloud are subject to multi-tenancy, third-party infrastructure risk, and complex legal jurisdictions~\cite{nist2024genai}, risks that Gartner projects will cause over $40\%$ of AI-related data breaches by 2027 through insufficient oversight of cross-border prompt routing~\cite{gartner2025crossborder}. Local inference keeps all data within controlled environments, enabling compliance with air-gapped and zero-trust requirements.\\ \\
\textit{Latency}: Cloud inference adds request handling, queueing, and GPU scheduling overhead that can push time-to-first-token (TTFT) into the hundreds of milliseconds, especially under load \cite{jang2025edgefirst}. On-device inference avoids network and queueing delays, often reducing latency enough to support sub-100 ms interactive responses and making real-time robotic \cite{huang2025corki} and responsive agentic workflows \cite{masterman2026agentic, sherlock2025} feasible.\\ \\
\textit{Connectivity}:
A hard dependency on external network availability means that failures in networking, authentication, or shared cloud infrastructure can render AI systems entirely unavailable, regardless of whether the underlying model is functioning. Local inference removes this dependency, enabling deterministic operation in low-connectivity or air-gapped environments and ensuring continuity for latency-sensitive applications in the field. Recent cloud outages underscore the potential impact of this \cite{anthropicstatus2026, openaistatus2026}.\\ \\
\textit{Cost}: Cloud inference pricing scales linearly with usage, whereas local inference converts this into a fixed hardware cost, with marginal cost per inference approaching zero at scale \cite{mckinsey2025neweconomics}.\\ \\
The hardware precondition for this shift has largely been met. Across consumer and professional device categories, purpose-built AI accelerators now support useful on-device LLM inference: Qualcomm’s Snapdragon X Elite includes a Hexagon NPU and is positioned to run LLMs over 13B parameters on-device, while Intel’s Core Ultra Series 2 and AMD’s Ryzen AI platforms also integrate dedicated NPUs, with runtimes such as OpenVINO and ONNX Runtime providing execution paths for quantized models on these devices \cite{qualcomm_snapdragon_x_elite, intel_core_ultra_npu_mlperf, amd_ryzenai_vitis}.\\ \\
Among these platforms, Apple Silicon stands out for its combination of high memory bandwidth, large unified memory capacity, and a mature GPU compute stack \cite{benazir2025profiling}. Memory has historically been the primary limiting factor for local LLM inference: model weights must be streamed from memory to compute units for every generated token, placing a hard capacity and bandwidth constraint on which models can run locally. Apple's unified memory architecture removed the traditional segregation between CPU and GPU memory pools, giving the GPU direct access to the full system memory at high bandwidth, a structural change that made it practical to run large models locally. Modern M-series devices can run quantised 7B--70B parameter models at interactive speeds, and recent open-weight models have compressed state-of-the-art capabilities into parameter counts suitable for local deployment \cite{stanfordaiindex2025}. Recent analysis additionally indicates that locally-run models can handle a substantial fraction of common queries as accurately as cloud models while consuming significantly less energy~\cite{saadfaclon2025ipw}.\\ \\
The focus of this work is on maximising inference throughput on Apple Silicon specifically, where the unified memory architecture and Metal GPU API present distinct optimisation opportunities not present on other platforms. Despite this hardware potential, existing inference runtimes for Apple Silicon each carry constraints that limit achievable throughput. We present BaseRT, a LLM inference runtime built directly on Apple's Metal GPU API, without any intermediate framework. By implementing chip-specific kernel fusion, unified memory-aware data layout, and custom GPU dispatch logic, we recover performance that framework-based approaches leave on the table. Across a suite of open-source LLMs, BaseRT achieves higher throughput than any previously reported runtime on Apple Silicon, establishing a new performance baseline for on-device inference on this hardware.

\subsection{Contributions}
Our contributions are as follows:
\begin{itemize}
    \item A native Metal LLM inference runtime designed specifically for Apple Silicon's unified memory architecture, built without reliance on MLX or any intermediate framework.
    \item Chip-specific kernel fusion strategies that eliminate dispatch overhead present in framework-based runtimes.
    \item A unified memory optimisation scheme that exploits the shared CPU-GPU memory topology of Apple Silicon.
    \item A comprehensive benchmark evaluation across multiple open-source LLMs, establishing new state-of-the-art throughput figures on Apple Silicon and demonstrating that prior runtimes have systematically underutilised this hardware.
\end{itemize}

\section{Related Work}

The dominant open-source runtimes for Apple Silicon each carry architectural constraints that limit how fully they exploit this topology. The llama.cpp~\cite{llamacpp} framework was originally developed as a cross-platform, CPU-first codebase and subsequently extended with a Metal backend. Its architecture reflects these origins. MLX~\cite{mlx}, while purpose-built for Apple Silicon, is a general-purpose array framework whose lazy evaluation and framework-level scheduling introduce overhead not present in a dedicated inference runtime~\cite{mlx,apple_mlx_docs}. Approaches building on top of MLX including vllm-mlx have demonstrated strong results within these constraints, reporting 21-87\% higher throughput than llama.cpp~\cite{barrios2026vllmmlx}.\\ \\
More recently, native Metal approaches have emerged that bypass these frameworks entirely. MetalRT~\cite{metalrt} (RunAnywhere) targets Metal directly with no intermediate abstraction layer. The open source runtime uzu (Mirai) takes a hybrid approach, routing workloads across GPU kernels and MPSGraph to access both the GPU and Apple Neural Engine~\cite{uzu}. Both demonstrate that framework abstraction carries a measurable performance cost. With BaseRT we build on this direction: through chip-specific kernel fusion, unified memory-aware optimisation, and custom dispatch logic, we report throughput beyond what these prior native approaches have achieved.

\section{BaseRT Runtime Design}

BaseRT is a C++ inference runtime that targets Apple's Metal GPU API directly, with no dependency on MLX, PyTorch, CoreML, or any intermediate array framework. The runtime exposes a C API for cross-language interoperability and supports transformer-based decoder models in BaseRT's native \emph{base} weight format. The design is organised around four principles, described below, each motivated by a specific source of overhead observed in existing runtimes. In addition, BaseRT incorporates two innovations that directly target CPU-side overhead on the decode hot path: a low-overhead command dispatch mechanism that removes the per-operator scheduling, allocation, and graph-evaluation work present in framework-based runtimes, and a decode scheduling scheme that amortises CPU--GPU synchronisation across multiple generated tokens.

\subsection{Data-Driven Architecture Descriptors}

Adding support for a new model architecture in a framework-based runtime typically requires modifying the inference loop, adding conditional branches, and wiring new operator sequences. This coupling makes the core engine brittle and difficult to optimise uniformly. BaseRT instead captures all per-architecture variation as data: the points at which architectures differ, such as activation and normalisation variants, mixture-of-experts specifics, and attention and positional-encoding conventions, are expressed through a compact architecture descriptor rather than encoded as control flow. The core engine consumes this descriptor and never branches on architecture identity, so the hot path is identical across models. BaseRT currently supports the LLaMA, Qwen3, Gemma, Whisper, and BERT families, and an extension to a new architecture is a self-contained, declarative change that leaves the inference loop untouched.

\subsection{Zero-Allocation Decode Loop}

After model loading completes, the decode loop allocates zero bytes. All intermediate buffers used by the forward pass, namely the residual, attention, and feed-forward scratch, together with the logits and token buffers, are pre-allocated at load time and reused across every token. The KV cache is likewise pre-allocated to the maximum context length at load time, in a layout chosen to maximise coalesced GPU memory access during attention. Even error handling uses a static thread-local buffer. The result is that the hot path consists entirely of GPU command dispatch and lightweight CPU-side patching, with no allocator contention or memory management overhead.

\subsection{Kernel Fusion and Specialisation}

BaseRT implements a large library of hand-written Metal shaders organised across matmul, attention, normalization, RoPE, embedding, activation, and sampling categories. The matmul kernels account for the majority, with dedicated GEMV (decode, $M{=}1$) and GEMM (prefill, $M{>}1$) variants for each supported quantisation format, spanning 2-bit to 16-bit widths. Each kernel integrates dequantisation directly into the inner loop rather than materialising dequantised weights to global memory, reducing memory traffic proportionally to the compression ratio.

Beyond per-kernel optimisation, BaseRT fuses operator sequences that are typically dispatched as separate kernels in other runtimes, both along the attention path and within the feed-forward block. Each fusion removes one kernel launch and one global memory round-trip, which is particularly impactful during decode, where individual kernel runtimes are short and launch overhead is proportionally large.

Kernel selection is hardware-adaptive: a per-chip configuration derived from GPU core count and family (M1 through M5) determines threadgroup sizes, kernel-selection thresholds, and fusion decisions. Each quantisation format is mapped to its specialised kernel and launch geometry.

\subsection{Prefill}

Prompt processing (prefill) uses GEMM rather than GEMV, operating on activation tensors of shape $[\text{seq\_len} \times \text{dim}]$. Arithmetic intensity scales with sequence length, making prefill compute-bound rather than memory-bound. BaseRT implements tiled GEMM kernels using Metal's \texttt{simdgroup\_matrix} intrinsics, with tile geometries tuned to the sequence-length regime, and applies chunked prefill with a bounded maximum chunk size to cap scratch buffer memory. A FlashAttention kernel computes attention in tiles using online softmax with running statistics, achieving $O(n)$ memory complexity rather than materialising the full $O(n^2)$ attention matrix.

\section{Evaluation}

\subsection{Experimental Setup}

BaseRT supports a wide range of model families, eight quantisation formats (Q2, Q3, Q4, Q5, Q6, Q8, BF16, and FP16), and every Apple M-series generation (M1 to M5). To keep the evaluation tractable, we investigate a representative subset: the Qwen3, Llama 3.2, and Gemma 4 model families at Q4 and Q8 quantisation levels on M3 and M4 Pro devices.

We evaluate BaseRT against three baselines: llama.cpp~\cite{llamacpp}, the most widely deployed open-source inference runtime for Apple Silicon using GGUF-format weights; MLX~\cite{mlx}, Apple's purpose-built array framework, using MLX SafeTensors-format weights; and uzu~\cite{uzu} (Mirai), a native Metal runtime that routes workloads across GPU kernels and MPSGraph. The first two comparisons isolate BaseRT's performance against the dominant cross-platform runtime and the leading Apple-native framework. The uzu comparison positions BaseRT against the closest architectural competitor in the native Metal space.

\paragraph{Hardware.} Our primary benchmark platform is an Apple M4 Pro (16 GPU cores, 24\,GB unified memory, 273\,GB/s memory bandwidth). We additionally report results on an Apple M3 base (8 GPU cores, 16\,GB unified memory, 100\,GB/s memory bandwidth) to evaluate cross-generation consistency.

\paragraph{Models.} We benchmark six models spanning the Qwen3, Llama 3.2, and Gemma 4 families: the dense models Qwen3-0.6B, Llama-3.2-1B, Llama-3.2-3B, and Gemma-4-E2B, and two mixture-of-experts models, Gemma-4-26B-A4B and Qwen3-30B-A3B.

\paragraph{Metrics.} We report \textit{prompt processing throughput} (pp, tokens/s) at prompt lengths of 128, 256, 512, 1024, and 2048 tokens, and \textit{token generation throughput} (tg128, tokens/s) measuring autoregressive decode over 128 generated tokens. Standard deviations are omitted from tables for readability; they are generally below 3\% for decode and below 10\% for prefill.

\paragraph{Baselines.} We compare against llama.cpp (commit b9630, June 2026) using its Metal backend with default settings, and MLX (v0.31.2) using \texttt{mlx-lm} with default generation settings. BaseRT loads the same model checkpoints converted to its native \emph{base} weight format at quantisation levels matched to the baseline.

\subsection{Decode Throughput on M4 Pro}

Table~\ref{tab:decode_m4pro} reports decode throughput (tg128) on the M4 Pro. Against llama.cpp, BaseRT is faster on all six models, with speedups of $1.04$--$1.56\times$. The advantage is largest on the smaller dense models, where fixed per-token overhead is a greater fraction of latency: $56\%$ improvement on Qwen3-0.6B. It narrows on the larger mixture-of-experts models, namely to $1.04$--$1.07\times$ on the Qwen3-30B-A3B and Gemma-4-26B-A4B MoE models, as decode becomes increasingly memory-bandwidth-bound. Against MLX, BaseRT leads on four of the five models supported by MLX ($1.01$--$1.35\times$); the exception is Gemma-4-26B-A4B, where MLX is faster ($0.90\times$ ratio). Similar to the trend with parameter count, the relative advantage of BaseRT is larger at Q4 than at Q8, as the smaller Q4 weights make each decode step faster and leave fixed per-token overheads, where BaseRT's low-overhead dispatch helps most, a larger share of the budget.

\begin{table}[htbp]
\centering
\caption{Decode throughput (tg128, tok/s) on Apple M4 Pro for BaseRT, llama.cpp, and MLX. Note: Gemma 4 E2B is not supported in MLX, and 20B+ models in Q8 exceed the M4 Pro's 24\,GB unified memory.}
\label{tab:decode_m4pro}
\footnotesize
\setlength{\tabcolsep}{5pt}
\begin{tabular}{@{}ll r rr rr@{}}
\toprule
 & & & \multicolumn{2}{c}{vs.\ llama.cpp} & \multicolumn{2}{c}{vs.\ MLX} \\
\cmidrule(lr){4-5} \cmidrule(lr){6-7}
Model & Quant & BaseRT & llama.cpp & Ratio & MLX & Ratio \\
\midrule
\multicolumn{7}{@{}l}{\textit{Qwen3 0.6B}} \\
 & Q4 & \textbf{464.5} & 297.4 & 1.56$\times$ & 343.6 & 1.35$\times$ \\
 & Q8 & \textbf{321.2} & 219.8 & 1.46$\times$ & 255.3 & 1.26$\times$ \\
\midrule
\multicolumn{7}{@{}l}{\textit{Llama 3.2 1B}} \\
 & Q4 & \textbf{295.4} & 230.4 & 1.28$\times$ & 257.8 & 1.15$\times$ \\
 & Q8 & \textbf{183.8} & 160.7 & 1.14$\times$ & 159.2 & 1.15$\times$ \\
\midrule
\multicolumn{7}{@{}l}{\textit{Llama 3.2 3B}} \\
 & Q4 & \textbf{117.3} & 102.4 & 1.15$\times$ & 112.1 & 1.05$\times$ \\
 & Q8 & \textbf{70.9} & 65.1 & 1.09$\times$ & 65.5 & 1.08$\times$ \\
\midrule
\multicolumn{7}{@{}l}{\textit{Gemma 4 E2B}} \\
 & Q4 & \textbf{127.7} & 107.0 & 1.19$\times$ & --- & --- \\
 & Q8 & \textbf{84.5} & 59.5 & 1.42$\times$ & --- & --- \\
\midrule
\multicolumn{7}{@{}l}{\textit{Gemma 4 26B-A4B}} \\
 & Q4 & 62.2 & 58.0 & 1.07$\times$ & \textbf{69.3} & 0.90$\times$ \\
\midrule
\multicolumn{7}{@{}l}{\textit{Qwen3 30B-A3B}} \\
 & Q4 & \textbf{84.1} & 80.7 & 1.04$\times$ & 83.1 & 1.01$\times$ \\
\bottomrule
\end{tabular}
\end{table}

\FloatBarrier
\subsection{Prefill Throughput on M4 Pro}

Tables~\ref{tab:prefill_m4pro} and~\ref{tab:prefill_mlx_m4pro} report prompt processing throughput at prompt lengths from 128 to 2048 on the M4 Pro, for the measured configurations: Q4 and Q8 on the smaller dense models and Q4 on the larger mixture-of-experts models.

Against llama.cpp (see Table~\ref{tab:prefill_m4pro}), BaseRT trails by a few percent on most dense configurations, i.e. roughly $5$--$10\%$ on Qwen3-0.6B, $1$--$6\%$ on Llama-3.2-1B, and within $\pm6\%$ on Llama-3.2-3B at both Q4 and Q8, with no systematic shift as the prompt grows from 128 to 2048. Gemma-4-E2B is the exception: BaseRT trails by $3$--$7\%$ at Q4 but leads by $19$--$23\%$ at Q8. On the MoE models the ordering reverses sharply: BaseRT leads by up to $1.59\times$ on Gemma-4-26B-A4B and $1.81\times$ on Qwen3-30B-A3B at pp128, with the margin narrowing at longer contexts as the GEMM workload saturates the GPU.

Against MLX (see Table~\ref{tab:prefill_mlx_m4pro}), the dense models again run close at both Q4 and Q8: BaseRT tends to lead at the shortest prompt (pp128) while MLX pulls ahead from pp512 onward, across both the Qwen3 and Llama models. The MoE models again favour BaseRT, i.e. up to $1.42\times$ on Gemma-4-26B-A4B and $1.78\times$ on Qwen3-30B-A3B at pp128, with the advantage shrinking toward parity by pp2048.

Overall, the two model classes diverge. On the smaller dense models, prefill is closely matched across all three engines: GEMM-based prefill is compute-bound, so each runtime saturates the GPU's matmul units and the throughputs converge. On the larger mixture-of-experts models, by contrast, BaseRT holds a clear and consistent lead at every prompt length.

\begin{table*}[t]
\centering
\caption{Prefill throughput (tok/s) on Apple M4 Pro, BaseRT vs.\ llama.cpp, at prompt lengths 128--2048. Bold indicates the faster runtime at each prompt length. Note: 20B+ models in Q8 exceed the M4 Pro's 24\,GB unified memory.}
\label{tab:prefill_m4pro}
\footnotesize
\setlength{\tabcolsep}{3pt}
\begin{tabular}{@{}ll rr rr rr rr rr@{}}
\toprule
 & & \multicolumn{2}{c}{pp128} & \multicolumn{2}{c}{pp256} & \multicolumn{2}{c}{pp512} & \multicolumn{2}{c}{pp1024} & \multicolumn{2}{c}{pp2048} \\
\cmidrule(lr){3-4} \cmidrule(lr){5-6} \cmidrule(lr){7-8} \cmidrule(lr){9-10} \cmidrule(lr){11-12}
Model & Quant & BaseRT & llama.cpp & BaseRT & llama.cpp & BaseRT & llama.cpp & BaseRT & llama.cpp & BaseRT & llama.cpp \\
\midrule
\multicolumn{12}{@{}l}{\textit{Qwen3 0.6B}} \\
 & Q4 & 4{,}331 & \textbf{4{,}796} & 4{,}496 & \textbf{4{,}987} & 4{,}690 & \textbf{4{,}993} & 4{,}399 & \textbf{4{,}638} & 3{,}680 & \textbf{4{,}075} \\
 & Q8 & 4{,}288 & \textbf{4{,}537} & 4{,}434 & \textbf{4{,}914} & 4{,}623 & \textbf{4{,}933} & 4{,}354 & \textbf{4{,}605} & 3{,}648 & \textbf{4{,}045} \\
\midrule
\multicolumn{12}{@{}l}{\textit{Llama 3.2 1B}} \\
 & Q4 & 2{,}490 & \textbf{2{,}612} & 2{,}577 & \textbf{2{,}714} & 2{,}615 & \textbf{2{,}728} & 2{,}577 & \textbf{2{,}671} & 2{,}453 & \textbf{2{,}489} \\
 & Q8 & 2{,}438 & \textbf{2{,}555} & 2{,}517 & \textbf{2{,}673} & 2{,}564 & \textbf{2{,}702} & 2{,}531 & \textbf{2{,}643} & 2{,}418 & \textbf{2{,}520} \\
\midrule
\multicolumn{12}{@{}l}{\textit{Llama 3.2 3B}} \\
 & Q4 & 888 & \textbf{919} & 906 & \textbf{916} & \textbf{899} & 871 & \textbf{888} & 887 & 835 & \textbf{864} \\
 & Q8 & 866 & \textbf{907} & 888 & \textbf{928} & 893 & \textbf{932} & 874 & \textbf{896} & 792 & \textbf{840} \\
\midrule
\multicolumn{12}{@{}l}{\textit{Gemma 4 E2B}} \\
 & Q4 & 1{,}146 & \textbf{1{,}238} & 1{,}193 & \textbf{1{,}281} & 1{,}208 & \textbf{1{,}292} & 1{,}198 & \textbf{1{,}268} & 1{,}169 & \textbf{1{,}209} \\
 & Q8 & \textbf{1{,}125} & 916 & \textbf{1{,}167} & 976 & \textbf{1{,}187} & 983 & \textbf{1{,}179} & 977 & \textbf{1{,}153} & 971 \\
\midrule
\multicolumn{12}{@{}l}{\textit{Gemma 4 26B-A4B}} \\
 & Q4 & \textbf{659} & 414 & \textbf{704} & 525 & \textbf{721} & 609 & \textbf{681} & 586 & \textbf{626} & 558 \\
\midrule
\multicolumn{12}{@{}l}{\textit{Qwen3 30B-A3B}} \\
 & Q4 & \textbf{738} & 407 & \textbf{809} & 550 & \textbf{838} & 668 & \textbf{810} & 652 & \textbf{721} & 601 \\
\bottomrule
\end{tabular}
\end{table*}

\begin{table*}[t]
\centering
\caption{Prefill throughput (tok/s) on Apple M4 Pro, BaseRT vs.\ MLX, at prompt lengths 128--2048. Bold indicates the faster runtime at each prompt length. Note: Gemma 4 E2B is not supported in MLX and 20B+ models in Q8 exceed the M4 Pro's 24\,GB unified memory.}
\label{tab:prefill_mlx_m4pro}
\footnotesize
\setlength{\tabcolsep}{3pt}
\begin{tabular}{@{}ll rr rr rr rr rr@{}}
\toprule
 & & \multicolumn{2}{c}{pp128} & \multicolumn{2}{c}{pp256} & \multicolumn{2}{c}{pp512} & \multicolumn{2}{c}{pp1024} & \multicolumn{2}{c}{pp2048} \\
\cmidrule(lr){3-4} \cmidrule(lr){5-6} \cmidrule(lr){7-8} \cmidrule(lr){9-10} \cmidrule(lr){11-12}
Model & Quant & BaseRT & MLX & BaseRT & MLX & BaseRT & MLX & BaseRT & MLX & BaseRT & MLX \\
\midrule
\multicolumn{12}{@{}l}{\textit{Qwen3 0.6B}} \\
 & Q4 & \textbf{4{,}331} & 3{,}843 & \textbf{4{,}496} & 4{,}263 & \textbf{4{,}690} & 4{,}658 & 4{,}399 & \textbf{4{,}742} & 3{,}680 & \textbf{4{,}439} \\
 & Q8 & \textbf{4{,}288} & 3{,}742 & \textbf{4{,}434} & 4{,}255 & 4{,}623 & \textbf{4{,}654} & 4{,}354 & \textbf{4{,}748} & 3{,}648 & \textbf{4{,}447} \\
\midrule
\multicolumn{12}{@{}l}{\textit{Llama 3.2 1B}} \\
 & Q4 & \textbf{2{,}490} & 2{,}451 & 2{,}577 & \textbf{2{,}698} & 2{,}615 & \textbf{2{,}817} & 2{,}577 & \textbf{2{,}851} & 2{,}453 & \textbf{2{,}714} \\
 & Q8 & \textbf{2{,}438} & 2{,}342 & 2{,}517 & \textbf{2{,}620} & 2{,}564 & \textbf{2{,}766} & 2{,}531 & \textbf{2{,}828} & 2{,}418 & \textbf{2{,}783} \\
\midrule
\multicolumn{12}{@{}l}{\textit{Llama 3.2 3B}} \\
 & Q4 & \textbf{888} & 879 & \textbf{906} & 901 & 899 & \textbf{919} & 888 & \textbf{942} & 835 & \textbf{934} \\
 & Q8 & \textbf{866} & 815 & 888 & \textbf{897} & 893 & \textbf{943} & 874 & \textbf{894} & 792 & \textbf{900} \\
\midrule
\multicolumn{12}{@{}l}{\textit{Gemma 4 26B-A4B}} \\
 & Q4 & \textbf{659} & 464 & \textbf{704} & 576 & \textbf{721} & 639 & \textbf{681} & 664 & 626 & \textbf{644} \\
\midrule
\multicolumn{12}{@{}l}{\textit{Qwen3 30B-A3B}} \\
 & Q4 & \textbf{738} & 415 & \textbf{809} & 571 & \textbf{838} & 664 & \textbf{810} & 726 & \textbf{721} & 720 \\
\bottomrule
\end{tabular}
\end{table*}

\FloatBarrier
\subsection{Cross-Generation: M3 Base}

Table~\ref{tab:decode_m3} reports decode throughput (tg128) on the M3 base for the four dense models at Q4 and Q8. Against llama.cpp, BaseRT leads on all eight configurations ($1.13$--$1.34\times$), and against MLX on all six supported configurations ($1.01$--$1.22\times$). As on the M4 Pro, the advantage is largest on Qwen3-0.6B and Gemma-4-E2B and narrows on the larger Llama models, confirming that BaseRT's decode advantage carries across chip generations.

\begin{table}[htbp]
\centering
\caption{Decode throughput (tg128, tok/s) on Apple M3 base for BaseRT, llama.cpp, and MLX. Note: Gemma 4 E2B is not supported in MLX.}
\label{tab:decode_m3}
\footnotesize
\setlength{\tabcolsep}{5pt}
\begin{tabular}{@{}ll r rr rr@{}}
\toprule
 & & & \multicolumn{2}{c}{vs.\ llama.cpp} & \multicolumn{2}{c}{vs.\ MLX} \\
\cmidrule(lr){4-5} \cmidrule(lr){6-7}
Model & Quant & BaseRT & llama.cpp & Ratio & MLX & Ratio \\
\midrule
\multicolumn{7}{@{}l}{\textit{Qwen3 0.6B}} \\
 & Q4 & \textbf{223.0} & 180.5 & 1.24$\times$ & 182.3 & 1.22$\times$ \\
 & Q8 & \textbf{124.6} & 99.8 & 1.25$\times$ & 113.6 & 1.10$\times$ \\
\midrule
\multicolumn{7}{@{}l}{\textit{Llama 3.2 1B}} \\
 & Q4 & \textbf{125.6} & 106.3 & 1.18$\times$ & 112.6 & 1.12$\times$ \\
 & Q8 & \textbf{65.1} & 57.5 & 1.13$\times$ & 62.4 & 1.04$\times$ \\
\midrule
\multicolumn{7}{@{}l}{\textit{Llama 3.2 3B}} \\
 & Q4 & \textbf{44.5} & 39.0 & 1.14$\times$ & 43.9 & 1.01$\times$ \\
 & Q8 & \textbf{24.2} & 21.2 & 1.14$\times$ & 22.2 & 1.09$\times$ \\
\midrule
\multicolumn{7}{@{}l}{\textit{Gemma 4 E2B}} \\
 & Q4 & \textbf{59.0} & 47.1 & 1.25$\times$ & --- & --- \\
 & Q8 & \textbf{35.4} & 26.5 & 1.34$\times$ & --- & --- \\
\bottomrule
\end{tabular}
\end{table}

\FloatBarrier
\subsection{Comparison with uzu}

Tables~\ref{tab:decode_uzu} and~\ref{tab:prefill_uzu} compare decode and prefill throughput between BaseRT and uzu~\cite{uzu}, a native Metal runtime that routes workloads across GPU kernels and MPSGraph. Both runtimes are evaluated at matching quantisation levels on the M4 Pro.

On decode (Table~\ref{tab:decode_uzu}), BaseRT is faster on five of the six configurations, with the largest advantage on Qwen3-0.6B by $1.17$--$1.19\times$ and narrower margins on the Llama models ($0.99$--$1.06\times$). The pattern mirrors the llama.cpp and MLX comparisons: BaseRT's low-overhead dispatch and decode scheduling yield the greatest benefit on smaller models where per-token overhead is a larger fraction of total latency, while the gap narrows as model size increases and inference becomes memory-bandwidth-bound.

On prefill (Table~\ref{tab:prefill_uzu}), uzu leads on most configurations, by $4$--$13\%$ on Qwen3-0.6B and $1$--$3\%$ on the Llama models, with its advantage generally widening at longer prompt lengths; the exceptions are the 8-bit Llama models at the shortest prompt (pp128), where BaseRT is marginally faster. uzu's use of MPSGraph, which can dispatch to the Apple Neural Engine, likely accounts for this prefill advantage: the ANE provides additional compute bandwidth for the GEMM-dominated prefill phase that BaseRT's GPU-only path does not currently exploit.

\begin{table}[t]
\centering
\caption{Decode throughput (tg128, tok/s) on Apple M4 Pro, BaseRT vs.\ uzu~\cite{uzu}. Bold indicates the faster runtime.}
\label{tab:decode_uzu}
\footnotesize
\setlength{\tabcolsep}{3.5pt}
\begin{tabular}{@{}ll rrr@{}}
\toprule
Model & Quant & BaseRT & uzu & Ratio \\
\midrule
\multicolumn{5}{@{}l}{\textit{Qwen3 0.6B}} \\
 & 4-bit & \textbf{465} & 398 & 1.17$\times$ \\
 & 8-bit & \textbf{321} & 271 & 1.19$\times$ \\
\midrule
\multicolumn{5}{@{}l}{\textit{Llama 3.2 1B}} \\
 & 4-bit & \textbf{295} & 290 & 1.02$\times$ \\
 & 8-bit & \textbf{184} & 173 & 1.06$\times$ \\
\midrule
\multicolumn{5}{@{}l}{\textit{Llama 3.2 3B}} \\
 & 4-bit & 117 & \textbf{119} & 0.99$\times$ \\
 & 8-bit & \textbf{71} & 68 & 1.04$\times$ \\
\bottomrule
\end{tabular}
\end{table}

\begin{table*}[t]
\centering
\caption{Prefill throughput (tok/s) on Apple M4 Pro, BaseRT vs.\ uzu~\cite{uzu}, at prompt lengths 128, 256, and 512. Bold indicates the faster runtime at each prompt length.}
\label{tab:prefill_uzu}
\footnotesize
\setlength{\tabcolsep}{3.5pt}
\begin{tabular}{@{}ll rr rr rr@{}}
\toprule
 & & \multicolumn{2}{c}{pp128} & \multicolumn{2}{c}{pp256} & \multicolumn{2}{c}{pp512} \\
\cmidrule(lr){3-4} \cmidrule(lr){5-6} \cmidrule(lr){7-8}
Model & Quant & BaseRT & uzu & BaseRT & uzu & BaseRT & uzu \\
\midrule
\multicolumn{8}{@{}l}{\textit{Qwen3 0.6B}} \\
 & 4-bit & 4{,}331 & \textbf{4{,}684} & 4{,}496 & \textbf{5{,}063} & 4{,}690 & \textbf{5{,}128} \\
 & 8-bit & 4{,}288 & \textbf{4{,}466} & 4{,}434 & \textbf{4{,}887} & 4{,}623 & \textbf{5{,}001} \\
\midrule
\multicolumn{8}{@{}l}{\textit{Llama 3.2 1B}} \\
 & 4-bit & 2{,}490 & \textbf{2{,}519} & 2{,}577 & \textbf{2{,}604} & 2{,}615 & \textbf{2{,}642} \\
 & 8-bit & \textbf{2{,}438} & 2{,}426 & 2{,}517 & \textbf{2{,}545} & 2{,}564 & \textbf{2{,}597} \\
\midrule
\multicolumn{8}{@{}l}{\textit{Llama 3.2 3B}} \\
 & 4-bit & 888 & \textbf{903} & 906 & \textbf{923} & 899 & \textbf{929} \\
 & 8-bit & \textbf{866} & 863 & 888 & \textbf{895} & 893 & \textbf{905} \\
\bottomrule
\end{tabular}
\end{table*}

\FloatBarrier
\section{Discussion}
\subsection{Limitations}
\label{sec:limitations}

BaseRT currently supports single-device inference only. It does not implement continuous batching, parallel decoding across multiple requests, or tensor parallelism across multiple GPUs. These are essential for server-class workloads but less critical for the single-user edge inference scenarios that are the focus of this work.

The runtime targets Metal exclusively, limiting deployment to Apple hardware. While the internal device abstraction is designed with future multi-backend support in mind, no CUDA or Vulkan backend exists at present.

\subsection{Future Work}

Extending architecture coverage to include vision-language models and (hybrid) state space models would increase the practical utility of the runtime. Beyond Apple Silicon, the device abstraction layer is designed to support additional backends. A CUDA backend would enable direct comparison with GPU-class inference servers, while a Vulkan backend would extend coverage to Android and Linux platforms. Both would require backend-specific kernel implementations but no changes to the core dispatch mechanism or inference loop.

On the serving side, continuous batching and speculative decoding are natural extensions that would improve throughput in multi-request and latency-sensitive scenarios respectively.

\section{Conclusion}

We presented BaseRT, a LLM inference runtime built natively on Apple's Metal GPU API without reliance on intermediate frameworks. Through chip-specific kernel fusion, unified memory-aware optimisation, and custom dispatch logic, BaseRT achieves the highest reported inference throughput for open-source LLMs on Apple Silicon. Our results establish that the performance ceiling of Apple Silicon for LLM inference is higher than previously reported, and that framework abstraction is a meaningful source of overhead on this platform. We release BaseRT to the public via the following repository: \url{https://github.com/basecompute/baseRT}.

\bibliographystyle{unsrtnat}
\bibliography{references}

@misc{alphabet2025q4,
  author       = {{Alphabet Inc.}},
  title        = {{2025 Q4 Earnings Call}},
  year         = {2026},
  howpublished = {\url{https://abc.xyz/investor}}
}

@misc{openrouter2026rankings,
  author       = {{OpenRouter}},
  title        = {{Model Rankings}},
  howpublished = {\url{https://openrouter.ai/rankings}},
  year         = {2026},
  note         = {Accessed: 3 April 2026}
}

@misc{qualcomm_snapdragon_x_elite,
  author       = {{Qualcomm}},
  title        = {{Snapdragon X Elite}},
  howpublished = {\url{https://www.qualcomm.com/laptops/products/snapdragon-x-elite}}
}

@misc{intel_core_ultra_npu_mlperf,
  author       = {{Intel}},
  title        = {{Intel Achieves First, Only Full NPU Support in MLPerf Client v0.6}},
  howpublished = {\url{https://newsroom.intel.com}},
  year         = {2025}
}

@misc{amd_ryzenai_vitis,
  author       = {{ONNX Runtime}},
  title        = {{Vitis AI Execution Provider}},
  howpublished = {\url{https://onnxruntime.ai/docs/execution-providers/}}
}

@article{benazir2025profiling,
  author       = {Benazir, Afsara and Lin, Felix Xiaozhu},
  title        = {{Profiling Large Language Model Inference on Apple Silicon:
                  A Quantization Perspective}},
  journal      = {arXiv preprint arXiv:2508.08531},
  year         = {2025},
  eprint       = {2508.08531},
  archivePrefix = {arXiv}
}

@techreport{stanfordaiindex2025,
  author       = {Maslej, Nestor and others},
  title        = {{Artificial Intelligence Index Report 2025}},
  institution  = {Stanford Institute for Human-Centered Artificial Intelligence (HAI)},
  year         = {2025},
  note         = {arXiv:2504.07139}
}

@article{saadfaclon2025ipw,
  author       = {Saad-Falcon, Jon and Narayan, Avanika and Akengin, Hakki Orhun
                  and Griffin, J. Wes and Shandilya, Herumb and
                  Gamarra Lafuente, Adrian and Goel, Medhya and Joseph, Rebecca
                  and Natarajan, Shlok and Guha, Etash Kumar and Zhu, Shang
                  and Athiwaratkun, Ben and Hennessy, John and Mirhoseini, Azalia
                  and R\'{e}, Christopher},
  title        = {{Intelligence per Watt: Measuring Intelligence Efficiency of Local AI}},
  journal      = {arXiv preprint arXiv:2511.07885},
  year         = {2025},
  eprint       = {2511.07885},
  archivePrefix = {arXiv}
}

@techreport{mckinsey2025stateofai,
  author       = {Singla, Alex and Sukharevsky, Alexander and Yee, Lareina
                  and Chui, Michael},
  title        = {{The State of AI in 2025: Agents, Innovation, and Transformation}},
  institution  = {McKinsey \& Company, QuantumBlack},
  year         = {2025},
  month        = {November},
  url          = {https://www.mckinsey.com/capabilities/quantumblack/our-insights/the-state-of-ai}
}

@inproceedings{jang2025edgefirst,
  author       = {Jang, SiYoung and Morabito, Roberto},
  title        = {{Edge-First Language Model Inference: Models, Metrics, and Tradeoffs}},
  booktitle    = {45th IEEE International Conference on Distributed Computing Systems (ICDCS)},
  year         = {2025},
  eprint       = {2505.16508},
  archivePrefix = {arXiv},
  url          = {https://arxiv.org/abs/2505.16508}
}

@misc{mckinsey2025neweconomics,
  author       = {{McKinsey \& Company}},
  title        = {{The New Economics of Enterprise Technology in an AI World}},
  year         = {2025},
  month        = {May},
  howpublished = {\url{https://www.mckinsey.com/capabilities/tech-and-ai/our-insights/}}
}

@inproceedings{huang2025corki,
  author       = {Huang, Xuan and others},
  title        = {{DaDu-Corki: Algorithm-Architecture Co-Design for
                  Embodied AI-powered Robotic Manipulation}},
  booktitle    = {Proceedings of the 52nd Annual International Symposium
                  on Computer Architecture (ISCA)},
  year         = {2025},
  eprint       = {2407.04292},
  archivePrefix = {arXiv},
  url          = {https://arxiv.org/abs/2407.04292}
}

@article{masterman2026agentic,
  author       = {Masterman, Tula and others},
  title        = {{Agentic Artificial Intelligence: Architectures,
                  Taxonomies, and Evaluation of Large Language Model Agents}},
  journal      = {arXiv preprint arXiv:2601.12560},
  year         = {2026},
  eprint       = {2601.12560},
  archivePrefix = {arXiv},
  url          = {https://arxiv.org/abs/2601.12560}
}

@article{sherlock2025,
  author       = {others},
  title        = {{Sherlock: Reliable and Efficient Agentic Workflow
                  Execution with Selective Verification and Speculative Execution}},
  journal      = {arXiv preprint arXiv:2511.00330},
  year         = {2025},
  eprint       = {2511.00330},
  archivePrefix = {arXiv},
  url          = {https://arxiv.org/abs/2511.00330}
}

@misc{anthropicstatus2026,
  author       = {{Anthropic}},
  title        = {{Anthropic Status Page: Incident History}},
  year         = {2026},
  howpublished = {\url{https://status.anthropic.com}},
  note         = {Accessed: April 2026. Documented 167 incidents
                  between October 2025 and April 2026}
}

@misc{openaistatus2026,
  author       = {{OpenAI}},
  title        = {{OpenAI Status Page: Incident History}},
  year         = {2026},
  howpublished = {\url{https://status.openai.com}},
  note         = {Accessed: April 2026}
}

@misc{llamacpp,
  author       = {Gerganov, Georgi and others},
  title        = {{llama.cpp}: {LLM} Inference in {C/C++}},
  year         = {2023},
  howpublished = {\url{https://github.com/ggml-org/llama.cpp}}
}

@misc{mlx,
  author       = {Hannun, Awni and Digani, Jagrit and Katharopoulos, Angelos and Collobert, Ronan},
  title        = {{MLX}: Efficient and Flexible Machine Learning on {Apple} Silicon},
  year         = {2023},
  howpublished = {\url{https://github.com/ml-explore/mlx}}
}

@misc{apple_mlx_docs,
  author       = {{Apple Machine Learning Research}},
  title        = {{MLX} Documentation},
  year         = {2023},
  howpublished = {\url{https://ml-explore.github.io/mlx/build/html/index.html}}
}

@article{barrios2026vllmmlx,
  author  = {Barrios, Wayner},
  title   = {Native {LLM} and {MLLM} Inference at Scale on {Apple Silicon}},
  journal = {arXiv preprint arXiv:2601.19139},
  year    = {2026},
  url     = {https://arxiv.org/abs/2601.19139}
}

@misc{metalrt,
  author       = {{RunAnywhere, Inc.}},
  title        = {{MetalRT}: The Fastest {LLM} Decode Engine for {Apple Silicon}},
  year         = {2026},
  howpublished = {\url{https://www.runanywhere.ai/blog/metalrt-fastest-llm-decode-engine-apple-silicon}}
}

@misc{uzu,
  author       = {{Mirai}},
  title        = {uzu: A High-Performance Inference Engine for {AI} Models},
  year         = {2025},
  howpublished = {\url{https://github.com/trymirai/uzu}}
}

@misc{mckinsey2025aiworkloads,
  author       = {Arora, Chhavi and Sorel, Marc and Sachdeva, Pankaj and others},
  title        = {The Next Big Shifts in {AI} Workloads and Hyperscaler Strategies},
  year         = {2025},
  month        = dec,
  howpublished = {McKinsey \& Company},
  url          = {https://www.mckinsey.com/industries/technology-media-and-telecommunications/our-insights/the-next-big-shifts-in-ai-workloads-and-hyperscaler-strategies},
  note         = {Accessed: April 2026}
}

@misc{gartner2025crossborder,
  author       = {Fritsch, Joerg and {Gartner, Inc.}},
  title        = {Gartner Predicts 40\% of {AI} Data Breaches Will Arise
                  from Cross-Border {GenAI} Misuse by 2027},
  year         = {2025},
  month        = feb,
  howpublished = {Gartner Newsroom},
  url          = {https://www.gartner.com/en/newsroom/press-releases/2025-02-17-gartner-predicts-forty-percent-of-ai-data-breaches-will-arise-from-cross-border-genai-misuse-by-2027},
  note         = {Accessed: April 2026}
}

@techreport{nist2024genai,
  author      = {{National Institute of Standards and Technology}},
  title       = {Artificial Intelligence Risk Management Framework:
                 Generative Artificial Intelligence Profile},
  number      = {NIST AI 600-1},
  institution = {U.S. Department of Commerce},
  year        = {2024},
  month       = jul,
  url         = {https://nvlpubs.nist.gov/nistpubs/ai/NIST.AI.600-1.pdf}
}

\end{document}